%% file: main.tex
\pdfoutput=1

\documentclass[11pt]{article}

\usepackage[preprint]{acl}

\usepackage{times}
\usepackage{latexsym}
\usepackage{multirow}
\usepackage{booktabs}
\usepackage[T1]{fontenc}

\usepackage[utf8]{inputenc}

\usepackage{microtype}

\usepackage{inconsolata}

\usepackage{graphicx}
\usepackage{listings}
\usepackage{amssymb}
\usepackage{pifont}
\usepackage[frozencache,cachedir=.]{minted}
\usepackage{subcaption}
\usepackage{amsmath}

\title{Meaning Typed Prompting: A Technique for Efficient, Reliable Structured Output Generation}

\author{Chandra Irugalbandara \\
  ASCII Corp \\
  \texttt{irugalbandara@ascii.ai}}

\begin{document}
\maketitle
\begin{abstract}
Extending Large Language Models (LLMs) to advanced applications requires reliable structured output generation. Existing methods which often rely on rigid JSON schemas, can lead to unreliable outputs, diminished reasoning capabilities, and increased computational overhead, limiting LLMs' adaptability for complex tasks. We introduce \textbf{Meaning Typed Prompting (MTP)}, a technique for efficient structured output generation that integrates types, meanings, and abstractions, such as variables and classes, into the prompting process. By utilizing expressive type definitions, MTP enhances output clarity and reduces dependence on complex abstractions, simplifying development, and improving implementation efficiency. This enables LLMs to understand relationships and generate structured data more effectively. Empirical evaluations on multiple benchmarks demonstrate that MTP outperforms existing frameworks in accuracy, reliability, consistency, and token efficiency.  We present \textbf{Semantix}, a framework that implements MTP, providing practical insights into its application\footnote{Source codes for Semantix is available \href{https://github.com/chandralegend/semantix}{here}}.
\end{abstract}

\input{introduction}
\input{related_work}
\input{methodology}
\input{experimental_setup}
\input{results_n_analysis}
\input{conclusions}
\input{limitations}

\subsection{Ethics Statement}
We obtained the necessary permissions to use the dataset provided by \citep{ner_dataset, bastianelli-etal-2020-slurp, massive_amazon}. We have utilized AI assistants, specifically Grammarly and ChatGPT, to correct grammatical errors and rephrase sentences.

\bibliography{main}

\clearpage

\appendix
\input{experiments}
\input{case_studies}

\end{document}

%% file: introduction.tex
\section{Introduction}

\begin{figure*}[!h!t]
    \centering
    \includegraphics[width=1\linewidth]{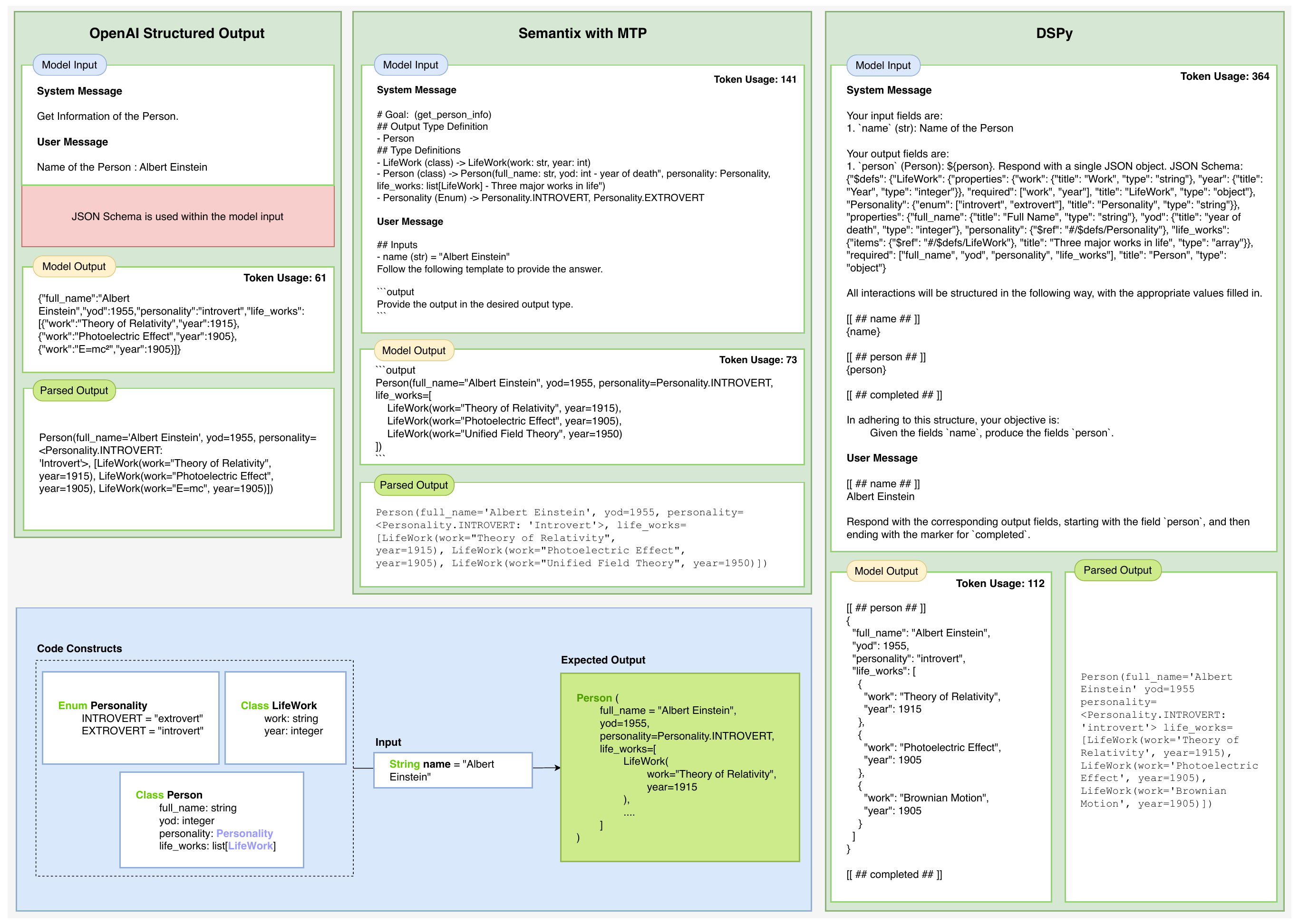}
    \caption{Comparison of structured output methods—OpenAI, Semantix, and DSPy—using identical code constructs and the same LLM (gpt-4o-mini). OpenAI and DSPy convert types into JSON Schema and output JSON; Semantix uses natural language representations and outputs object representations. Semantix has the lowest input token usage and, for longer outputs, consistently uses fewer tokens due to less frequent use of \{\}".}
    \label{fig:nested_structure_outputs}
\end{figure*}

Large Language Models (LLMs) have enabled the development of sophisticated applications, including complex task automation and domain-specific solutions \citep{xia2024fofobenchmarkevaluatellms} where strict adherence to formatting standards is crucial. Structured outputs improve AI integration into development tools by offering consistent output structures, simplifying error handling, and making LLM-generated responses more reliable for real-world use. With the rise of agentic frameworks \citep{autogen, agentinstruct}, having structured outputs is particularly important for seamless data parsing without an external output validator. 

Current approaches for generating structured outputs typically employ zero-shot \citep{zeroshot_reasoners} or few-shot prompting \citep{llm_fewshot_learner}, where developers provide instructions alongside JSON templates or Schemas to define the desired output format \citep{dspy, lmql, outline}. Post-processing techniques are then used to extract and validate the generated data. Although OpenAI’s constrained decoding strategy \citep{structured_output_api} ensures reliability, it limits reasoning capabilities due to its reliance on restriction on JSON Schemas \cite{let_me_speak_freely}. Additionally, detailed JSON Schemas increase token consumption and introduce challenges in maintaining syntactic correctness, often resulting in invalid outputs. Past attempts have been made to develop frameworks for structured output generation in LLMs. However, they either require full access to the language models \citep{outline}, use proprietary services such as Function Calling \citep{openai_function_2023, google_cloud_function_2023} or JSON Mode \citep{openai_json_2023} internally or suffer from issues of reliability and precision, frequently consuming excessive tokens \citep{instructor_github, modelsmith_github, llamaindex, dspy}. Furthermore, these methods often depend on hard-coded prompt templates or require manual prompting reducing adaptability. 

To address these limitations, we introduce Meaning Typed Prompting (MTP), a novel approach for structured output generation. Building on MTP, we present Semantix, a framework that uses MTP at its core to embed semantic information directly into type definitions with expressive meaning types, eliminating the need for additional abstractions. This approach reduces dependency on prompt configurations and simplifies the framework's learning curve. Inspired by MTLLM \cite{mtllm}, which introduces meaning-typed transformations, Semantix extends this paradigm by specifically targeting structured output generation through MTP. This advancement effectively addresses the limitations of JSON Schema based methods, function calling, and constrained decoding.

We evaluate Semantix on benchmarks for structured outputs, including multi-label classification, named entity recognition, and synthetic data generation \citep{structured_llm_benchmark}. To assess its impact on reasoning capabilities, we also conduct evaluations on GSM8k \citep{gsm8k} and MMMU \citep{mmmu}. Our empirical results demonstrate that Semantix significantly enhances output reliability, clarity, and token efficiency compared to existing methods. 

%% file: related_work.tex
\section{Related works}

\begin{table*}[!h!t]
\centering
\scriptsize
\begin{tabular}{@{}lcccccccc@{}}
\toprule
Feature & OpenAI & Instructor & LlamaIndex & Marvin & ModelSmith & DSPy & Fructose & Semantix (Ours) \\
\midrule
LLM-Agnostic & \ding{55} & \ding{55} & \checkmark & \ding{55} & \checkmark & \checkmark & \ding{55} & \checkmark \\
No use of Additional Abstractions & \ding{55} &  \ding{55} &  \ding{55} &  \ding{55} & \ding{55} & \ding{55} & \checkmark & \checkmark \\
Automated Output Fixing & \ding{55} & \checkmark & \ding{55} & \checkmark & \checkmark & \checkmark & \checkmark & \checkmark \\
No Manual Prompting Needed & \ding{55} & \ding{55} & \ding{55} & \checkmark & \ding{55} & \checkmark & \ding{55} & \checkmark \\
No use of JSON Mode / Function Calling & \ding{55} & \ding{55} & \ding{55} & \ding{55} & \checkmark & \checkmark & \ding{55} & \checkmark \\
Transparency & \ding{55} & \ding{55} & \ding{55} & \checkmark & \checkmark  & \checkmark & \checkmark & \checkmark \\
Vision Support & \checkmark & \checkmark & \checkmark & \checkmark & \ding{55} & \ding{55} & \ding{55} & \checkmark \\
\bottomrule
\end{tabular}
\caption{Comparison of frameworks based on various capabilities including model-agnosticism, additional abstractions, vision support, automated output fixing, manual prompting, transparency, and usage of JSON mode or function calling.}
\label{table:framework_comparison_features_switched}
\end{table*}

\begin{figure*}[!ht]
    \centering
    \includegraphics[width=1\linewidth]{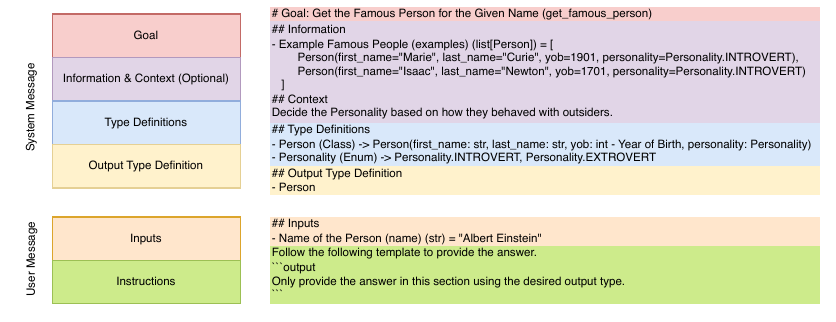}
    \caption{Structure of the Meaning Typed Prompt and example Final Prompt for generating a Person object for a given name with attributes as first name, last name, year of birth and personality as an enumerated type.}
    \label{fig:mtp_structure}
\end{figure*}

Structured output generation from LLMs is essential for ensuring consistency and efficiency \citep{building_your_own_copilot, llm_for_se}. Several frameworks tackle this challenge using different approaches. OpenAI’s structured outputs \citep{structured_output_api} employ JSON Schema-based constrained decoding to ensure outputs align with predefined formats. However, this method is restricted to closed-source APIs, limiting transparency, flexibility, and adaptability in diverse environments. Outline \citep{outline} extends this concept to open-source models by incorporating regular expressions, making it more accessible for developers.

Marvin \citep{marvin_github}, ModelSmith \citep{modelsmith_github}, Instructor \citep{instructor_github}, and Fructose \citep{fructose_github} rely on JSON schemas, leveraging JSON Mode or function-calling APIs, which are typically available only in closed-source LLM APIs. Fructose offers a simplified interface for these APIs but lacks a novel structural approach. DSPy \citep{dspy} also employs JSON schemas, operating independently of JSON Mode or function-calling APIs, and uses Pydantic models with post-generation assertions to correct output inconsistencies. However, the reliance on JSON schemas and Pydantic models increases development complexity and token consumption across these frameworks.

LMQL \citep{lmql} introduces a querying language that mimics manual prompting through iterative LLM calls \citep{modular_task_solution} to achieve structured outputs. While effective, this approach places an additional workload on developers, requiring careful prompt engineering and management of multiple queries, which can reduce flexibility in complex reasoning tasks. 

MTLLM \citep{mtllm} enhances structured output by embedding meaning into types through a construct called \textit{SemString}. This construct allows the attachment of meaningful explanations to types, variables, and functions. However, MTLLM relies on the Jac-lang domain-specific language and compile-time registries, introducing significant overhead and limiting accessibility.

In contrast, Semantix embeds semantic information directly into type definitions at runtime, eliminating the need for JSON Schemas, function calling, or domain-specific languages as shown in Table \ref{table:framework_comparison_features_switched}. This approach simplifies development, reduces token consumption, and ensures compatibility with a broad range of LLMs, offering a more flexible, transparent, and efficient solution.

%% file: methodology.tex
\section{Meaning Typed Prompting} \label{section:mtp}

We present Meaning Type Prompting (MTP), At its core we extend traditional built-in types and custom classes to expressive type definitions written in natural language. Moreover, we opt out from using popular JSON schema for structured output requests by using Pythonic class representation. Along with the expressive type definitions and Pythonic class representation, our final prompt consists of Goal, Information \& Context, Output Type Definition, Inputs, and Instructions as shown in Figure \ref{fig:mtp_structure}. 

To clarify the components of the MTP final prompt, we use the example of generating a person's information using a language model, as shown in Figure \ref{fig:mtp_structure}.

\subsection{Pythonic Class Representation}

Instead of relying on JSON Schema for defining output format like other frameworks, we leverage the LLM's capability to generate structured outputs directly using Pythonic class representations. By moving away from JSON Schema for hierarchical data representation, we omit the need for excessive use of special characters like \{, \}, ", '. This increases the token efficiency, especially for large outputs, and from our empirical results, it enhances reasoning capabilities.

For instance, when requesting a person's information, with attributes \texttt{first\_name (str)}, \texttt{last\_name (str)}, \texttt{yob (int)}, and \texttt{likes (list[str])}, the JSON Schema can be defined as:

\begin{minted}[tabsize=2,breaklines,fontsize=\footnotesize]{js}
{
  "properties":{
    "first_name":{"title": "First Name", "type": "string"},
    "last_name":{"title": "Last Name", "type": "string"},
    "yob":{"title": "Yob", "type": "string"},
    "likes":{
      "items":{
        "type": "string"
      },
      "title": "Likes", "type": "array"
    }
  },
  "required":["first_name", "last_name", "yob", "likes"],
  "title": "Person",
  "type": "object"
}
\end{minted}

In contrast, the equivalent Pythonic class representation in MTP is much more concise:

\begin{minted}[tabsize=2,breaklines,fontsize=\footnotesize, breaksymbolleft=, breaksymbolright=]{python}
Person (Class) -> Person(first_name: str, last_name: str, yob: int, likes: list[str])
\end{minted}

This approach significantly reduces overhead while improving both clarity and interpretability. This class representation with the types is added to the final prompt under the Type Definitions section of the system prompt (Figure \ref{fig:mtp_structure}).

\subsection{Expressive Type Definitions}

Expressive type definitions enhance traditional built-in types by adding context through natural language. For example, the \verb|yob: int| can be extended to \verb|yob: int - Year of Birth| to provide more clarity to the LLM. While using descriptive attribute names such as \verb|year_of_birth| instead of \verb|yob| is possible, it still may be insufficient for the LLM to grasp the context. Moreover, from the programmer's perspective, the extra-long identifiers can cause cluttered codes. To address this, the earlier Person class definition can be extended as follows:

\begin{minted}[tabsize=2,breaklines,fontsize=\footnotesize, breaksymbolleft=, breaksymbolright=]{python}
Person (Class) -> Person(first_name: str, last_name: str, yob: int - Year of Birth, likes: list[str])
\end{minted}

From our observations, the advantages of expressive type definitions are twofold: Defining types with natural languages helps the LLM to understand the context of the output requested reducing ambiguity. Also, it allows the LLM to constrain the output generation more precisely to the intended type. 

\subsection{Goal}

While explicitly defining a Goal is optional, it can be used to define the intended task for the LLM clearly and more descriptively. If not provided, the function name will be automatically used as the Goal. 

For example, an explicit goal will appear in the final prompt as follows:

\begin{minted}[tabsize=2,breaklines,fontsize=\footnotesize]{markdown}
Get the Famous Person for the Given Name
\end{minted}

\subsection{Information \& Context}

\textbf{Information} and \textbf{Context} optionally provide additional data to assist the LLM in generating accurate outputs. 

\textbf{Information} includes task-related elements or examples that guide the LLM in producing correctly formatted outputs which can be used to simulate few-shot prompting \citep{llm_fewshot_learner}. For example, additional information provided by the user will appear in the system prompt as follows:

\begin{minted}[tabsize=2,breaklines,fontsize=\footnotesize]{python}
Wikipedia Information (wiki_info) = "Albert Einstein was a German-born theoretical physicist..."

Example Famous People (examples) (list[Person]) = [
    Person(first_name="Marie", last_name="Curie", yob=1901, likes=["Research", ...]),
    Person(first_name="Isaac", last_name="Newton", yob=1701, likes=["Gravity", ...])
]
\end{minted}

\textbf{Context} provides simplified, static instructions for output generation. For example, if only the person’s likes outside their profession are needed the context can be specified as follows in the final prompt:

\begin{minted}[tabsize=2,breaklines,fontsize=\footnotesize, breaksymbolleft=, breaksymbolright=]{markdown}
Only consider their life outside their profession to identify likes
\end{minted}

\subsection{Output Type Definition}

This section specifies the function’s output type hint. For example, to generate structured data about a famous person, the output type can be defined as:

\begin{minted}[tabsize=2,breaklines,fontsize=\footnotesize]{markdown}
Person
\end{minted}

With prior type definitions, LLM is capable of recognizing and generating outputs in the extended format. We have also observed that the LLM can accurately generate outputs even when the output type hint is extended to nested types, such as \texttt{list[Person]}, \texttt{int}, \texttt{str}, or \texttt{tuple[Person, str]}.

\subsection{Inputs}

Inputs are the dynamic fields of a task. For example, when generating structured data about a famous person, the input could be their \texttt{name}. Inputs may vary in type (e.g., \texttt{str}, \texttt{int}) and quantity, with some tasks requiring multiple inputs or none. They can also carry meanings to guide generation. In MTP, inputs are defined as:

\begin{minted}[linenos,tabsize=2,breaklines,fontsize=\footnotesize, breaksymbolleft=, breaksymbolright=]{python}
Name of the Person (name) (str) = "Albert Einstein"
\end{minted}

\subsection{Instructions}

Instructions are directives for generating the output. In MTP, we guide the LLM to produce results within a Markdown code block (\verb|```|). Additional prompting methods, such as chain-of-thought reasoning, and reflexion can be used to enhance output quality.

\subsection{Final Prompt}

The Final Prompt is a combination of Goal, Type Definitions, Information and Context, Output Type Definition, Inputs, and Instructions in Markdown format. The prompt is intended to be used with the chat message style, including system messages and user messages. However, due to the structured nature of the prompt, it is also usable in a standard text completion setting.

\section{Semantix}

\begin{figure}
    \centering
    \includegraphics[width=1\linewidth]{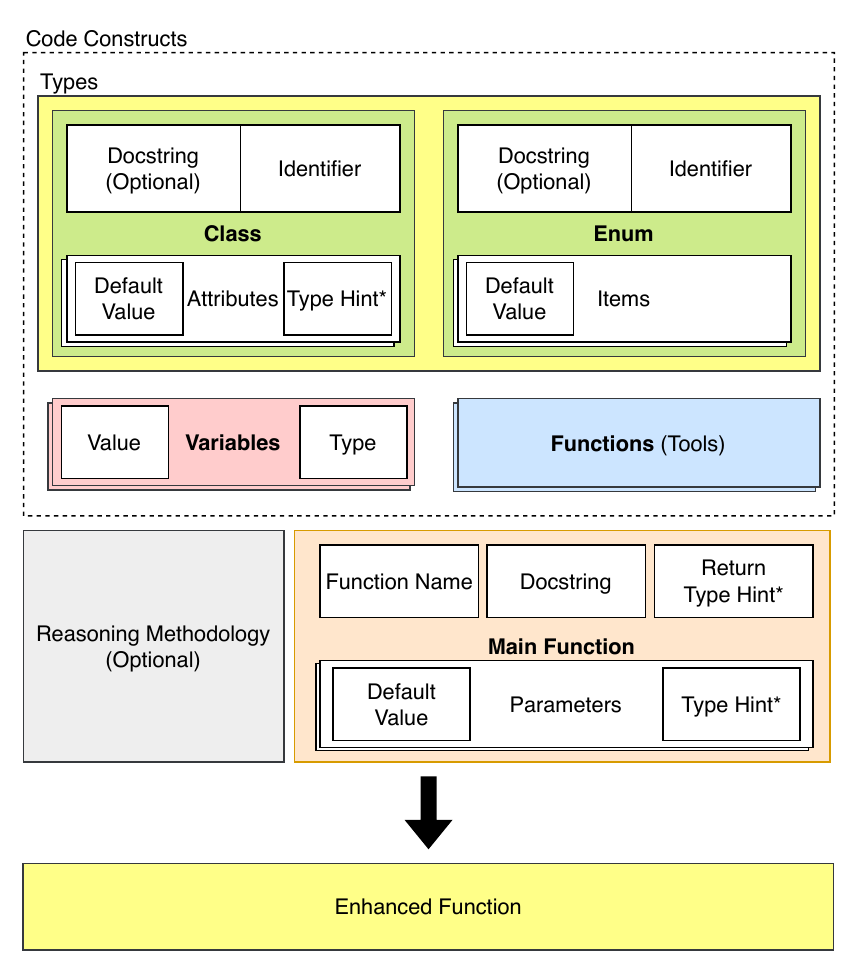}
    \caption{Semantix merges Types, Variables, Functions, and the Main Function into an enhanced function which generates MTP at runtime. *Type Hints can be extended using semantic types.}
    \label{fig:function_enhancement}
\end{figure}

\begin{figure}[t]
    \centering
    \includegraphics[width=1\linewidth]{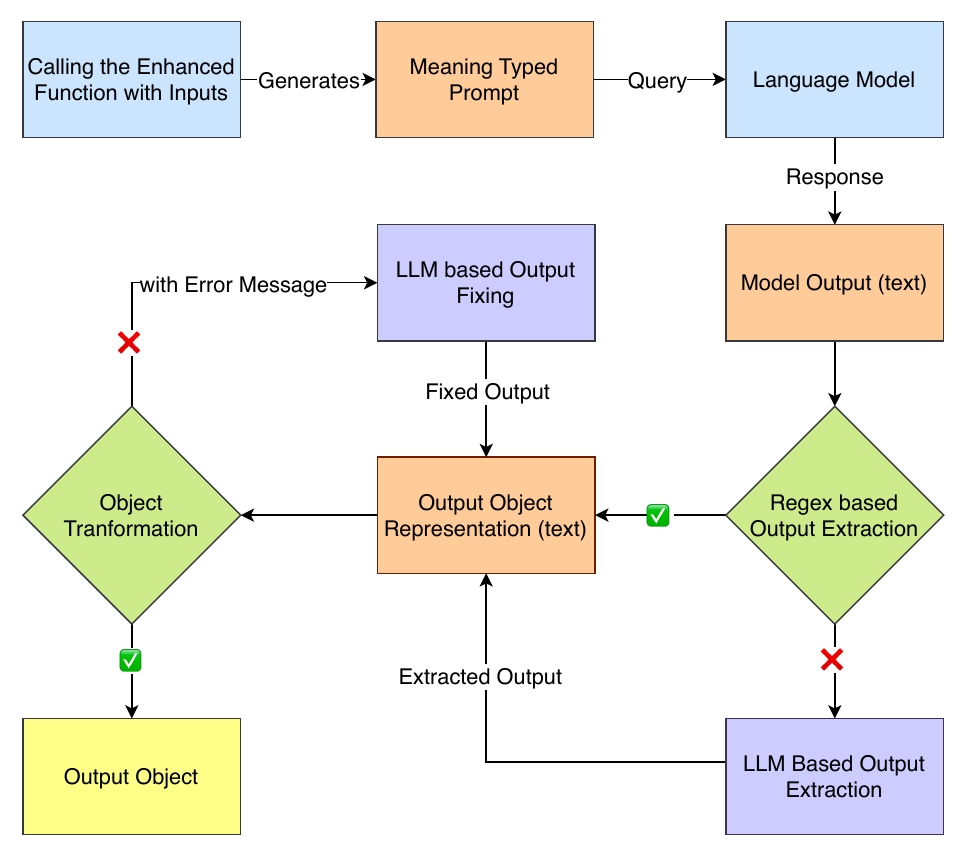}
    \caption{Execution of an Enhanced Function in Semantix includes querying the LLM, transforming the output into code, and managing errors.}
    \label{fig:execution}
\end{figure}

Semantix is a Python library that automates the generation of meaning-typed prompts. Semantix allows developers to enhance simple functions to enhanced functions simply by adding a decorator named \textit{enhance}. \textit{Semantic} type hint allows developers to add expressive type definitions. 

\paragraph{Semantic type hint.} Semantic is an optional type hint provided by Semantix to add expressive type definitions to a variable, argument of a class, or a parameter of a class in the MTP. Our observations indicate that explicit type definitions help the LLM generate outputs in the intended type or unit more reliably. For instance, when adding temperatures x and y where x is in Fahrenheit and y is in Celsius, and the result expected in Fahrenheit, Semantic type hints can be specified as follows.

\begin{minted}[tabsize=2,breaklines,fontsize=\footnotesize]{python}
# traditional function
def add(x:float, y: float) -> float:
    y_in_fahrenheit = (y * 9/5) + 32
    result = x + y_in_fahrenheit
    return result
\end{minted}

\begin{minted}[tabsize=2,breaklines,fontsize=\footnotesize]{python}
# with Semantic Type Hints
def add(x: Semantic[float, "Farenheit"], y: [float, "Celcius"]) -> Semantic[float, "Farenheit"]: ...
\end{minted}

As shown, Semantic types enhance clarity without changing the function name or arguments, ensuring backward compatibility while making the function easier to understand.

\paragraph{Function Enhancements.} Semantix offers a decorator referred to as ``enhance'', which can be employed with any function or method. Enhance can be used to specify the above-explained Goal, Information \& Context, or the Instruction method as shown in Figure \ref{fig:function_enhancement}.

\begin{minted}[tabsize=2,breaklines, fontsize=\footnotesize]{python}
@llm.enhance("Add the two temperatures accordingly", method="Chain-of-Thought")
def add(x: Semantic[float, "Farenheit"], y: [float, "Celcius"]) -> Semantic[float, "Farenheit"]: ...
\end{minted}



\paragraph{Execution.} During execution, Semantix dynamically generates a Meaning Typed Final Prompt to query the language model, retrieve the output, and transform it into the desired result (see Figure~\ref{fig:execution}). The \textit{Object Transformation Process} consists of two parts:

\paragraph{Output Extraction.} The prompt instructs the model to return output within a markdown block labeled "output". Although regular expressions typically extract the content, deviations from the expected format trigger a fallback to LLM-based extraction, where the model regenerates the output to ensure accuracy.

\paragraph{Object Transformation.} The extracted output is then processed using Python's Abstract Syntax Tree (AST) within the function's context to return the final object. In case of errors, such as syntax issues or unmatched brackets, the LLM is employed to debug and regenerate the output, improving reliability through iterative correction.

%% file: experimental_setup.tex
\section{Experiments}

\begin{table*}[!ht]
\scriptsize
\centering
\begin{tabular}{@{}lccc|ccc|ccc@{}}
\toprule
\multirow{2}{*}{Framework} & \multicolumn{3}{c|}{Multi-label Classification} & \multicolumn{3}{c|}{Named Entity Recognition} & \multicolumn{3}{c}{Synthetic Data Generation} \\
 & 0 retries & 2 retries & Consistency & 0 retries & 2 retries & Consistency & 0 retries & 2 retries & Consistency \\
\midrule
OpenAI & 0.658 & \underline{0.659} & \underline{0.998} & 0.801 & 0.773 & 0.965 & 0.783 & \textbf{0.928} & 0.814 \\
Instructor & 0.546 & 0.549 & 0.996 & \underline{0.837} & 0.825 & 0.986 & 0.803 & 0.874 & 0.912 \\
LlamaIndex & \underline{0.630} & 0.627 & 0.996 & 0.740 & 0.558 & 0.754 & 0.311 & 0.272 & 0.876 \\
Marvin & 0.000 & 0.000 & NaN & 0.617 & 0.606 & 0.982 & 0.082 & 0.082 & 1.000 \\
ModelSmith & 0.599 & 0.604 & 0.993 & 0.000 & 0.000 & NaN & 0.000 & 0.000 & NaN \\
Fructose & 0.537 & 0.537 & \textbf{1.000} & 0.826 & \underline{0.826} &\textbf{ 1.000} & \textbf{0.909} & \underline{0.923} & \underline{0.985} \\
Semantix (Ours) & \textbf{0.680} & \textbf{0.682} & \underline{0.998} & \textbf{0.842} & \textbf{0.843} & \underline{0.998} & \underline{0.902} & 0.902 & \textbf{1.000} \\
\bottomrule
\end{tabular}
\caption{GMS Scores and Consistency for Multilabel Classification, Named Entity Recognition, and Synthetic Data Generation Tasks (0 and 2 retries). For the framework reporting the maximum token usage on each benchmark, GMS is recorded as 0, resulting in Consistency being a NaN value, as defined.}
\label{table:gms_consistency}
\end{table*}

\begin{figure*}[t]
    \centering
    \begin{subfigure}[t]{0.32\textwidth}
        \centering
        \includegraphics[width=\linewidth]{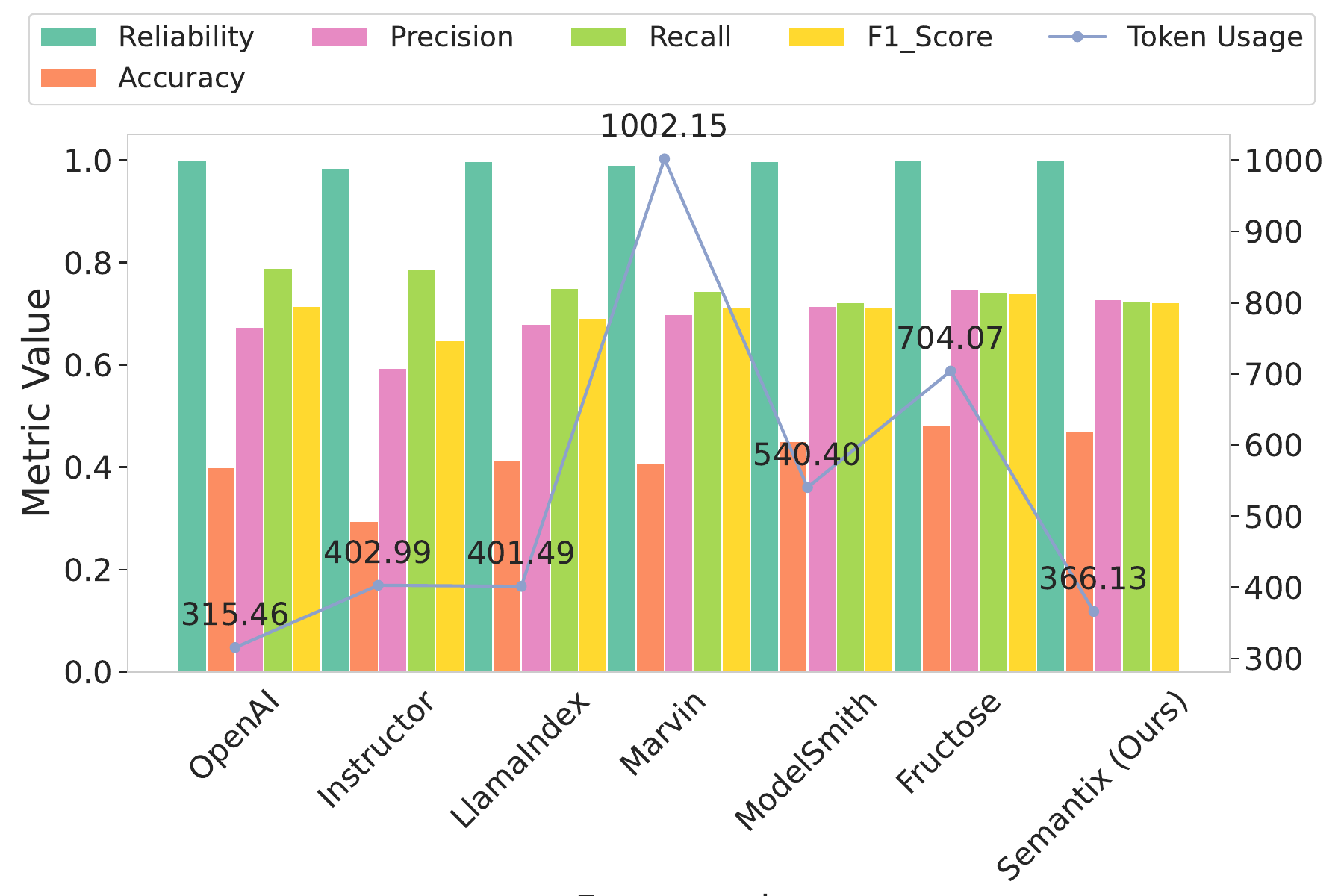}
        \caption{Multi-label Classification}
        \label{fig:multilabel_performance}
    \end{subfigure}\hfill
    \begin{subfigure}[t]{0.32\textwidth}
        \centering
        \includegraphics[width=\linewidth]{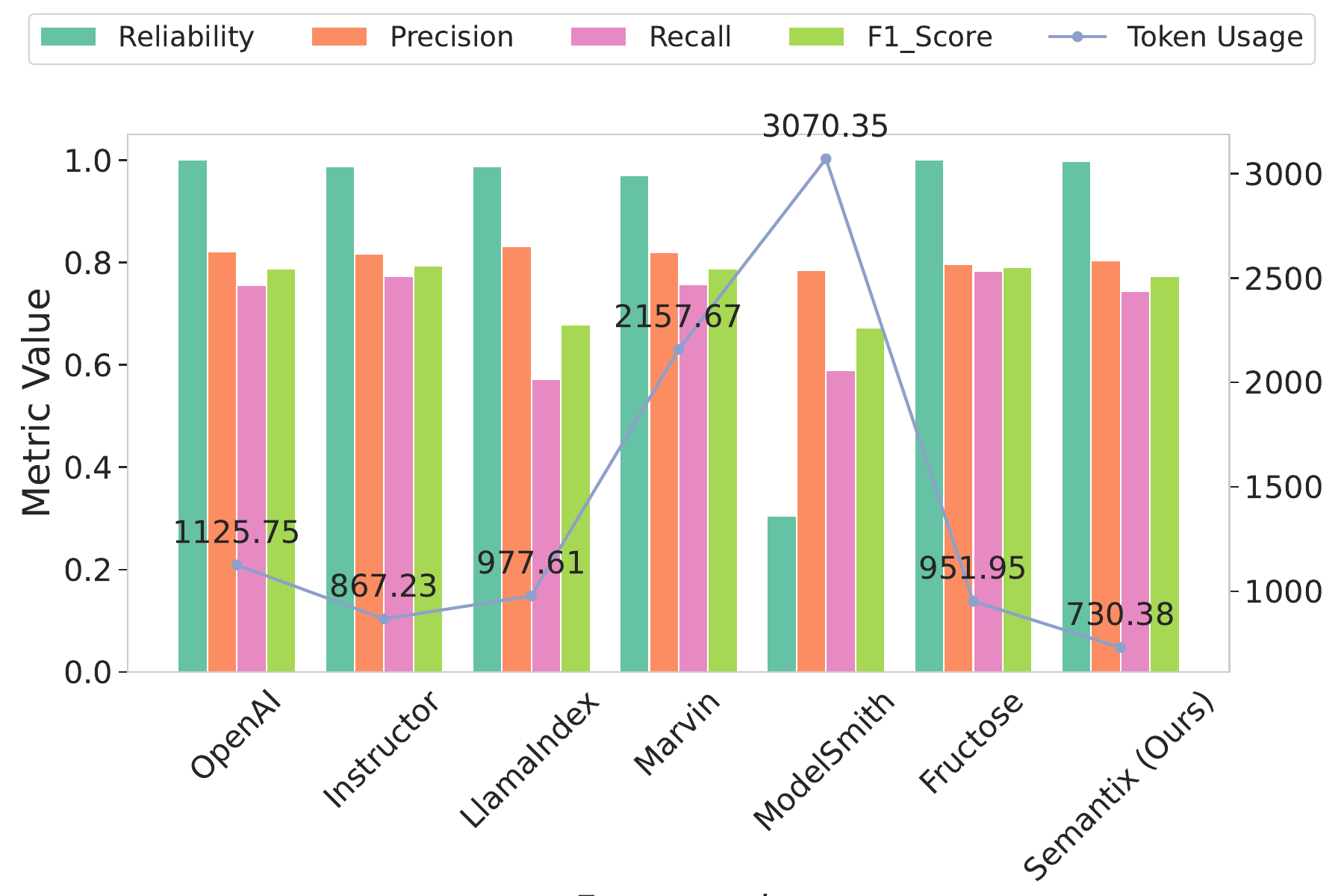}
        \caption{Named Entity Recognition}
        \label{fig:ner_performance}
    \end{subfigure}\hfill
    \begin{subfigure}[t]{0.32\textwidth}
        \centering
        \includegraphics[width=\linewidth]{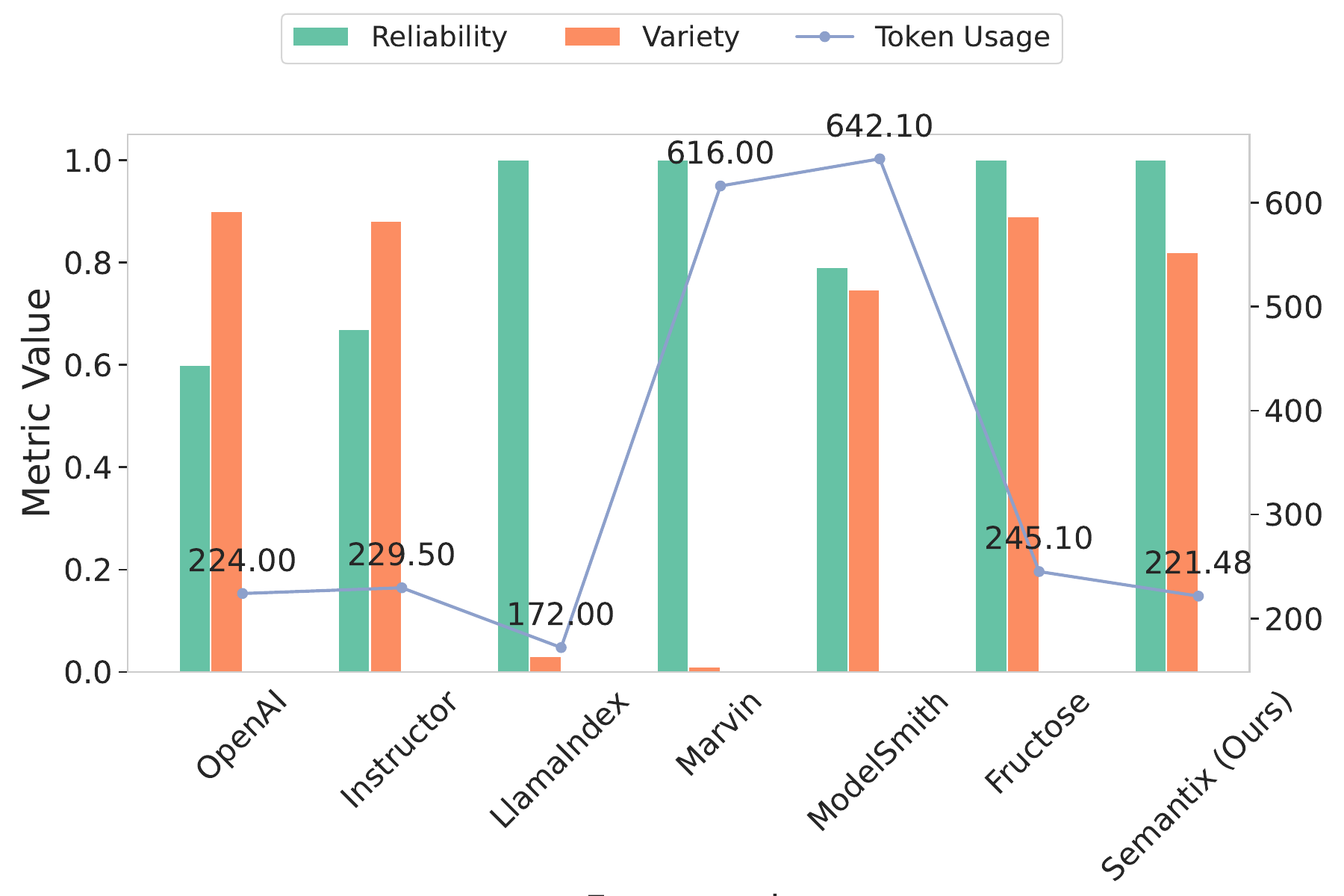}
        \caption{Synthetic Data Generation}
        \label{fig:synthetic_performance}
    \end{subfigure}
    \caption{Reliability and performance metrics and token usage for structured output generation benchmarks with 0 retries. Both Semantix and Fructose performed exceptionally well across all three benchmark tasks, but Semantix demonstrated significantly lower token usage.}
    \label{fig:combined_performance}
\end{figure*}

\subsection{Experimental setup}
This study evaluates Semantix in comparison to a variety of open-source frameworks, including Instructor, Modelsmith, Llamaindex, and Marvin, with OpenAI Structured Output as the closed-source benchmark. Each framework is assessed with retries $\in \{0, 2\}$, where ``retries'' signifies the number of supplementary attempts permitted if the framework does not succeed on the initial trial. All experiments were executed using GPT-4o-mini \citep{openai2024gpt4omini, openai2024gpt4technicalreport} under consistent conditions to ensure the equitable assessment of system-specific metrics such as latency.

\subsection{Benchmarks} 
We assess Semantix utilizing the LLM Structured Output Benchmarks \citep{structured_llm_benchmark}, which encompass Multi-label Classification, Named Entity Recognition (NER), and Synthetic Data Generation. \textbf{Multi-label Classification} applies the Alexa intent detection dataset \citep{massive_amazon, bastianelli-etal-2020-slurp}, requiring the framework to predict the labels associated with a given text. The text exemplifies a natural language command (e.g., wake me up at nine am on Friday), with the ground truth label (e.g., ``alarm set'') representing the intent of the command selected from a set of 60 labels. The synthetic test set, developed to evaluate the framework following \citep{structured_llm_benchmark}, consolidates multiple commands and necessitates the intent labels for all commands to be addressed in a single framework request. \textbf{Named-entity Recognition} is predicated on the Synthetic PII Finance Dataset \citep{ner_dataset}, whereby the framework must identify all entities corresponding to a set of labels for a specific text. \textbf{Synthetic Data Generation} is dedicated to the generation of fictitious personal information constrained to a strict format. More details on the exact required output format, dataset examples and sample outputs can be found at Appendix \ref{benchmark:multilabel}.

\subsection{Evaluation Metrics} 
We present an analysis of Reliability and Token Usage for all aforementioned experiments. Reliability is quantified as the percentage of instances where the framework successfully returns valid labels without errors in the exact required format as described in \citep{structured_llm_benchmark}. Token Usage is the total of prompt and completion tokens per query. For multi-label classification and named entity recognition tasks, micro-precision, micro-recall, and micro-F1 scores are reported, with provisions for partial credit in cases where the output is partially correct. Furthermore, we introduce the metric of exact accuracy, which assesses the degree to which the output entirely corresponds to the ground truth. We utilize the metric of variety to denote the percentage of unique names relative to the entirety of names generated by the framework during synthetic data generation experiments.

Given the obvious challenge of directly comparing these diverse metrics across different frameworks, to evaluate the performance of each framework, we calculate two main metrics: the \textbf{Geometric Mean Score (GMS)} and \textbf{Consistency}. These metrics provide a balanced view of reliability and performance across different tasks and trials. The GMS is computed by taking the geometric mean of the reliability, performance (F1 score or accuracy depending on the task), and normalized token usage (NTU). It is formulated as:

\small
\[
    \text{GMS} = \sqrt[3]{ \text{Reliability} \times \text{Performance} \times \text{NTU} }
\]
\normalsize

Where performance is typically represented by the F1 score or accuracy for a given task, and normalized token usage (NTU) is calculated as:

\small
\[
    \text{NTU} = 1 - \frac{\text{Token Usage} - \text{Min Token Usage}}{\text{Max Token Usage} - \text{Min Token Usage}}
\]
\normalsize

The Consistency metric quantifies the stability of the framework's performance across retries. It is defined as the relative difference between the GMS for 0 retries and 2 retries, computed as:

\small
\[
    \text{Consistency} = 1 - \left| \frac{\text{GMS}_{2 \text{ retries}} - \text{GMS}_{0 \text{ retries}}}{\text{GMS}_{0 \text{ retries}}} \right|
\]
\normalsize

A consistency score close to 1 indicates stable performance across retries, while a lower score signifies variability.

%% file: results_n_analysis.tex
\section{Results and Analysis}

We evaluated various frameworks across three tasks: Multi-label Classification, Named Entity Recognition (NER), and Synthetic Data Generation. The GMS and consistency metrics are shown in Table~\ref{table:gms_consistency}, with detailed results provided in Figure~\ref{fig:combined_performance} and Tables~\ref{table:framework_comparison_multilabel_combined}, \ref{table:framework_comparison_combined_ner}, and \ref{table:framework_comparison_combined_synthetic_datagen}.

\subsection{Multi-label Classification}

Semantix achieved the highest GMS scores for both retries = 0 and 2, demonstrating its ability to provide correct labels in the requested format while using only half the tokens compared to Fructose. All frameworks, except Marvin, reached perfect reliability when the number of retries increased.

\subsection{Named Entity Recognition Task}

As shown in Figure~\ref{fig:combined_performance}(b), Semantix, Fructose, and OpenAI achieved perfect reliability with retries. Instructor and Semantix had the highest precision for retries = 0 and 2, respectively, reflecting their effectiveness in entity identification. Semantix reported the best GMS score and the lowest token usage among all frameworks.

\subsection{Synthetic Data Generation Task}

In the synthetic data generation task (Figure~\ref{fig:combined_performance}(c)), most frameworks achieved high reliability. Fructose excelled in generating diverse data, as indicated by its high variety score. Semantix delivered comparable performance across all metrics with high reliability and consistency.

\subsection{Analysis}

We highlight several key findings from our evaluation:

\paragraph{Semantix reports the lowest token usage.} Meaning Typed Prompting (MTP) enables more concise and semantically rich prompts, leading to lower token usage compared to frameworks relying on JSON schemas or complex templates. This advantage is evident in tasks requiring larger outputs, such as NER, where Semantix reported the lowest token usage while achieving top performance.

\paragraph{Reliability improves with retries.} Most frameworks increased their reliability in generating correctly formatted outputs when the number of retries rose from 0 to 2. Semantix achieved perfect or near-perfect reliability across all tasks, even without retries, suggesting that MTP provides clearer instructions to the language model, reducing invalid outputs.

\paragraph{Generating synthetic data with a wide variety.} Fructose reported the highest variety in the synthetic data generation task and the best GMS for retries = 0, making it the leading framework for generating new data. Nevertheless, Semantix offered comparable results with high consistency.

\paragraph{Vision Support.} Utilizing type annotations, Semantix can accommodate various input data types such as text, images, and videos. This capability allows applications to preprocess inputs into formats suitable for the LLM. For instance, OpenAI's vision LLMs require images to be converted into base64 strings, while videos should be transformed into sequences of base64-encoded images. For qualitative results, refer to Appendix~\ref{food-analyzer}.

\paragraph{Semantix excels in reasoning tasks.} We evaluated the frameworks on the GSM8K and MMMU benchmarks to assess the impact of the language model's reasoning abilities (Table~\ref{table:gsm8k_mmmu_comparison}). GSM8K comprises mathematical problems in natural language, with outputs as integers or sets of integers. MMMU is a visual question-answering benchmark evaluated in the Accounting section. We compared OpenAI Structured Outputs, Fructose, and Semantix; Fructose lacks vision support and could not be evaluated on MMMU.

Semantix, when combined with a natural language to format conversion (NL-to-Format) step, outperformed other frameworks in both benchmarks. In NL-to-Format \cite{let_me_speak_freely}, we separate the reasoning process from the format generation process, allowing greater reasoning ability as it is not constrained by a specific output format . In standard Semantix, OpenAI, and Fructose, we use a class containing an attribute to hold reasoning steps. OpenAI Structured Output and Fructose use JSON mode with constrained decoding and cannot perform NL-to-Format strategy.

\begin{table}[!h]
\scriptsize
\centering
\begin{tabular}{@{}lcc@{}}
\toprule
Method & GSM8K & MMMU (Accounting) \\
\midrule
OpenAI Structured Outputs & 91.11 & 53.33 \\
Fructose & 93.49 & NA \\
Semantix & 91.95 & 60.00 \\
Semantix with NL-to-Format & \textbf{94.02} & \textbf{63.33} \\
\bottomrule
\end{tabular}
\caption{Comparison of GSM8K and MMMU (Accounting) accuracy using GPT-4o-mini for OpenAI Structured Outputs, Fructose, Semantix, and Semantix with NL-to-Format. Fructose lacks vision support, making MMMU evaluation impossible.}
\label{table:gsm8k_mmmu_comparison}
\end{table}

%% file: conclusions.tex
\section{Conclusion}

We introduced Meaning Typed Prompting (MTP), the core of Semantix—a framework that enhances structured output generation by embedding semantic types directly into prompts. Semantix offers a flexible alternative to JSON schema methods, surpassing them in token efficiency, reasoning capabilities, and output reliability. Notably, even without constrained decoding or models specifically trained for MTP, our approach performs exceptionally well, demonstrating that MTP's object representation is a viable alternative to JSON outputs. Our evaluations show that Semantix outperforms existing frameworks in tasks like multi-label classification, named entity recognition, and synthetic data generation while maintaining lower token usage and high consistency. With the implementation of constrained decoding for MTP and fine-tuning LLMs specifically for MTP, we anticipate outperforming JSON-based methods in every category. Future work will expand MTP to diverse LLM architectures, refine constrained decoding for better performance, and explore lightweight deployments for scalability and practical applications.


%% file: limitations.tex
\section{Limitations}

Semantix has several limitations to consider. First, it requires explicit type hints even for simple tasks, increasing the initial setup effort. However, this overhead diminishes in the projects where well-structured outputs are essential.

Second, it is still susceptible to hallucinations like all other frameworks. Since Semantix relies on standard prompting techniques without constrained decoding, LLMs may occasionally produce hallucinated outputs. While MTP shows potential for improved structured generation, addressing this issue remains a goal for future research.

Third, our testing has been limited to GPT-4o-mini. While we are confident that it works with other closed-source models like GPT-4, GPT-4o, Claude, and Gemini 1.5 Pro, we have not yet tested it with open-source models. Future work will explore MTP's performance with open-source LLMs such as LLaMA, Mistral, and Gemma.

Lastly, there is limited runtime type checking. As Python lacks native runtime type enforcement, Semantix currently offers limited type handling beyond developer-defined constructs. While Pydantic provides runtime validation, our framework aims to support greater flexibility. We are exploring the use of \texttt{pytypes} to enhance type checking and ensure more reliable outputs.

%% file: experiments.tex
\section{Experiments}
\subsection{Benchmark}
\subsubsection{Structured Output Generation Benchmark}

In this benchmark, several frameworks were compared across tasks requiring structured outputs, namely Multi-Label Classification and Named Entity Recognition (NER). The benchmark author synthesized two datasets for Multi-Label Classification by combining multiple natural language queries with corresponding intents to create samples with multiple labels. For NER, the author structured the ground truth by manipulating annotations from a PII Entity Recognition dataset. We used these datasets exactly as the author synthesized them.

The original benchmark evaluated frameworks including Fructose, ModelSmith, OpenAI Structured Output, Instructor, Outline, LMFormatEnforcer, LlamaIndex, Marvin, and Mirascope. We selected Fructose, ModelSmith, OpenAI Structured Outputs, Instructor, LlamaIndex, and Marvin for our evaluation. Mirascope was excluded as it was the worst-performing framework according to the initial benchmarks. We also did not consider Outline and LMFormatEnforcer because they require access to the LLM's decoding mechanism, necessitating the use of an open-source LLM, which is incompatible with closed-source LLMs accessed via APIs. However, we plan to evaluate Semantix with Outline and other constrained decoding techniques in future work.

\paragraph{Multilabel Classification} \label{benchmark:multilabel} In this benchmark task, we need to classify the query between 49 labels for a given natural language query. One query can have multiple Labels. This also tests the ability of the framework to handle Enumerators (Enums), where each label is an enum item. The expected output is a list of enum items related to the query. As you can see in the Code \ref{code:multilabel_classes} we have iteratively created an Enum Class Called \textit{Label} which will hold all the multilabel classes.

\begin{figure}[h]
\centering
\includegraphics[width=1\linewidth]{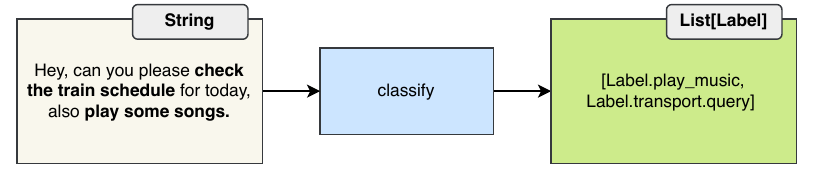} 
\caption{Multi-label Classification Example}
\label{img:classify}
\end{figure}

\begin{listing}
\begin{minted}[tabsize=2,breaklines,frame=lines,framesep=2mm,fontsize=\footnotesize]{python}
multilabel_classes = ['lists_createoradd', 'calendar_query', 'email_sendemail', 'news_query', 'play_music', 'play_radio', 'qa_maths', 'email_query', 'weather_query', 'calendar_set', 'iot_hue_lightdim', 'takeaway_query', 'social_post', 'email_querycontact', 'qa_factoid', 'calendar_remove', 'cooking_recipe', 'lists_query', 'general_quirky', 'alarm_query', 'takeaway_order', 'iot_hue_lightup', 'lists_remove', 'qa_currency', 'play_game', 'play_audiobook', 'qa_definition', 'music_query', 'datetime_query', 'transport_query', 'iot_hue_lightoff', 'iot_hue_lightchange', 'iot_hue_lighton', 'alarm_set', 'music_likeness', 'recommendation_movies', 'transport_ticket', 'recommendation_locations', 'audio_volume_mute', 'iot_wemo_on', 'play_podcasts', 'datetime_convert', 'audio_volume_other', 'recommendation_events', 'alarm_remove', 'iot_coffee', 'music_dislikeness', 'general_joke', 'social_query']

Label = create_enum("Label", {name: name for name in multilabel_classes}, "Multilabel Classes")

class MultiLabelClassification(BaseModel):
    classes: list[Label]
\end{minted}
\caption{Multi\-label Classes for the Multi\-label Classification} 
\label{code:multilabel_classes}
\end{listing}
\begin{listing}
\begin{minted}[tabsize=2,breaklines,frame=lines,framesep=2mm,fontsize=\footnotesize]{python}
@llm.enhance
def classify(text: str) -> list[Label]:...
\end{minted}
\caption{Semantix Implementation for Multilabel Classfication} 
\label{code:multilabel_semantix}
\end{listing}
\begin{listing}
\begin{minted}[tabsize=2,breaklines,frame=lines,framesep=2mm,fontsize=\footnotesize]{python}
def classify(text:str) -> list[Label]:
    response = openai_client.beta.chat.completions.parse(
        model='gpt-4o-mini',
        response_format=MultiLabelClassification,
        messages=[
            {"role": "user", "content": f"Classify the following text: {text}"}
        ],
    )
    return response.choices[0].message.parsed.classes
\end{minted}
\caption{OpenAI Structured Outputs Implementation for Multilabel Classification. Needs an additional Class to be made as everything should Pydantic Base models, such that it can be converted to a JSON Schema.} 
\label{code:multilabel_openai_instructor}
\end{listing}
\begin{listing}
\begin{minted}[tabsize=2,breaklines,frame=lines,framesep=2mm,fontsize=\footnotesize]{python}
@fructose_client
def classify(text: str) -> list[Label]:
    """Classify the given text."""
    ...
\end{minted}
\caption{Fructose Implementation for Multilabel Classification. Fructose has a similar look to semantix. However, you are required to perform manual prompting always, though your function names are descriptive enough.} 
\label{code:multilabel_fructose}
\end{listing}
\begin{listing}
\begin{minted}[tabsize=2,breaklines,frame=lines,framesep=2mm,fontsize=\footnotesize]{python}
text = "text to classify"

llamaindex_client = OpenAIPydanticProgram.from_defaults(
    output_cls= MultiLabelClassification,
    prompt_template_str="Classify the following text: {text}",
    llm_model= 'gpt-4o-mini,
)
output = llamaindex_client(text=text, description="Data model of items present in the text")

marvin.settings.openai.chat.completions.model = 'gpt-4o-mini'
output = marvin.cast(f"Classify the following text: {text}", MultiLabelClassification)

forge = Forge(
    model=OpenAIModel('gpt-4o-mini),
    response_model=MultiLabelClassification
)
output = forge.generate(user_input=f"Classify the following text: {text}")

# Have to do an additional step to get the expected output
output = output.classes
\end{minted}
\caption{Marvin, LlamaIndex, ModelSmith has a very similar implementation for Multilabel Classfication.} 
\label{code:multilabel_marvin_llamaindex_modelsmith}
\end{listing}

Every framework except semantix and fructose has to go through the extra step of creating a new pydantic base model and when results are available have to retrieve the expected output from the result. For all except semantix, the user has to manually provide a prompt where semantix is fully synthesized using function signature, function name, and type hints.

\paragraph{Named Entity Recognition (NER)} \label{benchmark:ner} In this task, the objective is to identify phrases (entities) in the given text that get classified into a given set of classes. This checks the ability to extract entities from long documents and also the ability to follow Optional Types (places where None is possible)

\begin{figure}[!h]
\centering
\includegraphics[width=\linewidth]{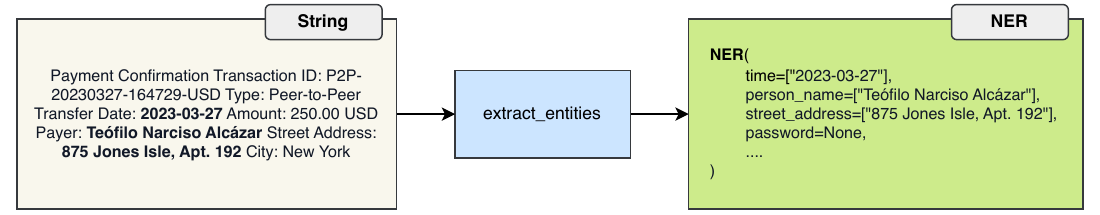} 
\caption{Named Entity Recognition (NER) Example}
\label{img:extract_entities}
\end{figure}

\begin{listing}
\begin{minted}[tabsize=2,breaklines,frame=lines,framesep=2mm,fontsize=\footnotesize]{python}
ner_entities = ['passport_number', 'bank_routing_number', 'account_pin', 'swift_bic_code', 'password', 'credit_card_number', 'email', 'phone_number', 'person_name', 'iban', 'ipv6', 'api_key', 'street_address', 'company', 'local_latlng', 'time', 'employee_id', 'customer_id', 'date_of_birth', 'ipv4', 'bban']

fields = {name: (Optional[list[str]], None) for name in ner_entities}

NER = create_class("NER", **fields)
\end{minted}
\caption{Named Entities and Expected Output Type for the Named Entity Recognition} 
\label{code:ner_entities}
\end{listing}
\begin{listing}
\begin{minted}[tabsize=2,breaklines,frame=lines,framesep=2mm,fontsize=\footnotesize]{python}
@llm.enhance
def extract_entities(text: str) -> NER: ...
\end{minted}
\caption{Semantix Implementation for Named Entity Recognition} 
\label{code:ner_semantix}
\end{listing}
\begin{listing}
\begin{minted}[tabsize=2,breaklines,frame=lines,framesep=2mm,fontsize=\footnotesize]{python}
def extract_entities(text:str) -> NER:
    response = openai_client.beta.chat.completions.parse(
        model='gpt-4o-mini',
        response_format=NER,
        messages=[
            {"role": "user", "content": f"Extract and resolve a list of entities from the following text: {text}"}
        ],
    )
    return response.choices[0].message.parsed
\end{minted}
\caption{OpenAI Structured Outputs Implementation for NER. Instructor uses a similar implementation where 'openai\_client.beta` is replaced with `instructor\_client`.} 
\label{code:ner_openai_instructor}
\end{listing}
\begin{listing}
\begin{minted}[tabsize=2,breaklines,frame=lines,framesep=2mm,fontsize=\footnotesize]{python}
@fructose_client
def extract_entities(text: str) -> NER:
    """Extract named entities from the given text."""
    ...
\end{minted}
\caption{Fructose Implementation for Named Entity Recognition} 
\label{code:ner_fructose}
\end{listing}
\begin{listing}
\begin{minted}[tabsize=2,breaklines,frame=lines,framesep=2mm,fontsize=\footnotesize]{python}
text = "text to extract entities from"

llamaindex_client = OpenAIPydanticProgram.from_defaults(
    output_cls= NER,
    prompt_template_str=f"Extract and resolve a list of entities from the following text: {text}",
    llm_model= 'gpt-4o-mini,
)
output = llamaindex_client(text=text, description="Data model of items present in the text")

marvin.settings.openai.chat.completions.model = 'gpt-4o-mini'
output = marvin.cast(f"Extract and resolve a list of entities from the following text: {text}", NER)

forge = Forge(
    model=OpenAIModel('gpt-4o-mini),
    response_model=NER
)
output = forge.generate(user_input=f"Extract and resolve a list of entities from the following text: {text}")
\end{minted}
\caption{Marvin, LlamaIndex, ModelSmith has a very similar implementation for NER.} 
\label{code:ner_marvin_llamaindex_modelsmith}
\end{listing}

\paragraph{Synthetic Data Generation} \label{benchmark:synthetic} In this task we test the ability of the frameworks to create synthetic data which is important when we want to train models, fine-tune models for domain-specific use cases, etc. The most important aspect we should focus on in this task is the variety of the generated data. As we are feeding the LLM the same input every time there is a high chance that the LLM will generate the same data, reducing the variety of the generations. Semantix was able to achieve 82\% variety where the prompt is left unchanged across the evaluation. When we added a dynamic element into the prompt, for example, a random integer between 0 to 1000, we saw a significant increase in variety, where it reached 98\% variety. But the issue with this approach is as we are adding a random element to the prompt we lose the consistency.

\begin{figure}[!h]
\centering
\includegraphics[width=0.8\linewidth]{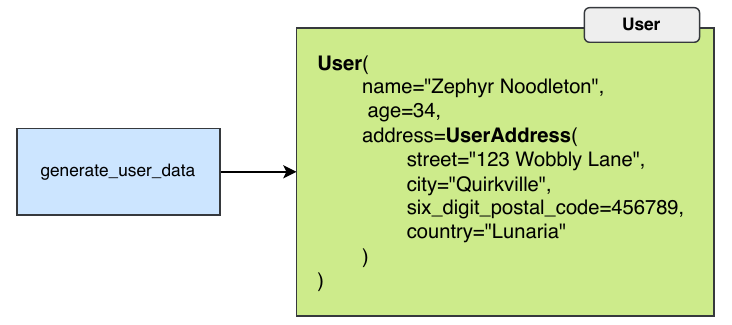} 
\caption{Synthetic Data Generation Example}
\label{img:generate_user_data}
\end{figure}

\begin{listing}
\begin{minted}[tabsize=2,breaklines,frame=lines,framesep=2mm,fontsize=\footnotesize]{python}

class UserAddress(BaseModel):
    street: str
    city: str
    six_digit_postal_code: int
    country: str

class User(BaseModel):
    name: str
    age: int
    address: UserAddress
\end{minted}
\caption{Named Entities and Expected Output Type for the Named Entity Recognition} 
\label{code:synthetic_user}
\end{listing}
\begin{listing}
\begin{minted}[tabsize=2,breaklines,frame=lines,framesep=2mm,fontsize=\footnotesize]{python}
@llm.enhance("Generate a random person's information. The name must be chosen at random. Make it something you wouldn't normally choose.")
def generate_user_data() -> User: ...
\end{minted}
\caption{Semantix Implementation for Synthetic Data Generation} 
\label{code:synthetic_semantix}
\end{listing}
\begin{listing}
\begin{minted}[tabsize=2,breaklines,frame=lines,framesep=2mm,fontsize=\footnotesize]{python}
def generate_user_data() -> User:
    response = openai_client.beta.chat.completions.parse(
        model='gpt-4o-mini',
        response_format=User,
        messages=[
            {"role": "user", "content": "generate a random person's information. The name must be chosen at random. Make it something you wouldn't normally choose."}
        ],
    )
    return response.choices[0].message.parsed
\end{minted}
\caption{OpenAI Structured Outputs Implementation for NER. Instructor uses a similar implementation where 'openai\_client.beta` is replaced with `instructor\_client`.} 
\label{code:syntheticl_openai_instructor}
\end{listing}
\begin{listing}
\begin{minted}[tabsize=2,breaklines,frame=lines,framesep=2mm,fontsize=\footnotesize]{python}
@fructose_client
def generate_user_data() -> User:
    """Generate a random person's information. The name must be chosen at random. Make it something you wouldn't normally choose."""
\end{minted}
\caption{Fructose Implementation for Named Entity} 
\label{code:synthetic_fructose}
\end{listing}
\begin{listing}
\begin{minted}[tabsize=2,breaklines,frame=lines,framesep=2mm,fontsize=\footnotesize]{python}
llamaindex_client = OpenAIPydanticProgram.from_defaults(
    output_cls= User,
    prompt_template_str="Generate a random person's information. The name must be chosen at random. Make it something you wouldn't normally choose.",
    llm_model= 'gpt-4o-mini,
)
output = llamaindex_client(text=text, description="Data model of items present in the text")

marvin.settings.openai.chat.completions.model = 'gpt-4o-mini'
output = marvin.cast("Generate a random person's information. The name must be chosen at random. Make it something you wouldn't normally choose.", NER)

forge = Forge(
    model=OpenAIModel('gpt-4o-mini),
    response_model=NER
)
output = forge.generate(user_input="Generate a random person's information. The name must be chosen at random. Make it something you wouldn't normally choose.")
\end{minted}
\caption{Marvin, LlamaIndex, ModelSmith has a very similar implementation for Synthetic Data Generation.} 
\label{code:synthetic_marvin_llamaindex_modelsmith}
\end{listing}

\begin{table*}[!ht]
\scriptsize
\centering
\begin{tabular}{@{}lcc|cc|cc|cc|cc|c@{}}
\toprule
\multirow{2}{*}{Framework} & \multicolumn{2}{c|}{Reliability} & \multicolumn{2}{c|}{Accuracy\footnote{Degree to which the output entirely corresponds the ground truth}} & \multicolumn{2}{c|}{Precision} & \multicolumn{2}{c|}{Recall} & \multicolumn{2}{c|}{F1 Score} & \multirow{2}{*}{Token Usage} \\
& 0 retries & 2 retries & 0 retries & 2 retries & 0 retries & 2 retries & 0 retries & 2 retries & 0 retries & 2 retries &  \\
\midrule
OpenAI & \textbf{1.000} & \textbf{1.000} & 0.399 & 0.401 & 0.674 & 0.675 & \textbf{0.789} & \textbf{0.791} & 0.714 & 0.714 & \textbf{315.46} \\
Instructor & 0.983 & \textbf{1.000} & 0.294 & 0.289 & 0.594 & 0.600 & 0.786 & 0.789 & 0.647 & 0.655 & 402.99 \\
LlamaIndex & 0.997 & \textbf{1.000} & 0.414 & 0.413 & 0.679 & 0.669 & 0.750 & 0.745 & 0.691 & 0.682 & 401.49 \\
Marvin & 0.990 & 0.991 & 0.409 & 0.401 & 0.699 & 0.699 & 0.744 & 0.752 & 0.711 & 0.713 & 1002.15 \\
ModelSmith & 0.998 & \textbf{1.000} & 0.450 & 0.457 & 0.715 & 0.713 & 0.722 & 0.729 & 0.713 & 0.716 & 540.40 \\
Fructose &\textbf{1.000} & \textbf{1.000} & \textbf{0.483} & \textbf{ 0.481} & \textbf{0.748} & \textbf{0.748} & 0.741 & 0.744 & \textbf{0.740} & \textbf{0.742} & 704.07 \\
Semantix (Ours) & \textbf{1.000} & \textbf{1.000} & \underline{0.471} & \underline{0.473} & \underline{0.728} & \underline{0.733} & 0.723 & 0.722 & \underline{0.722} & \underline{0.723} & \underline{366.13} \\
\bottomrule
\end{tabular}
\caption{Comparison of frameworks on the Multilabel Classification Task with 0 and 2 retries. Metrics include reliability, accuracy, precision, recall, F1 score, and token usage (reported for 2 retries). The table spans both columns for better readability without overflowing content.}
\label{table:framework_comparison_multilabel_combined}
\end{table*}

\begin{table*}[!ht]
\scriptsize
\centering
\begin{tabular}{@{}lcc|cc|cc|cc|c@{}}
\toprule
\multirow{2}{*}{Framework} & \multicolumn{2}{c|}{Reliability} & \multicolumn{2}{c|}{Precision} & \multicolumn{2}{c|}{Recall} & \multicolumn{2}{c|}{F1 Score} & \multirow{2}{*}{Token Usage} \\
& 0 retries & 2 retries & 0 retries & 2 retries & 0 retries & 2 retries & 0 retries & 2 retries &  \\
\midrule
OpenAI & \textbf{1.000} & \textbf{1.000} & 0.821 & 0.751 & 0.756 & 0.741 & 0.787 & 0.746 & 1125.75 \\
Instructor & 0.987 & 0.999 & 0.816 & 0.776 & \textbf{0.773} & \textbf{0.772} & \textbf{0.794} & 0.773 & 867.23 \\
LlamaIndex & 0.987 & 0.946 & \textbf{0.832} & 0.791 & 0.572 & 0.317 & 0.678 & 0.453 & 977.61 \\
Marvin & 0.970 & 0.960 & 0.819 & 0.802 & 0.757 & 0.741 & 0.787 & 0.770 & 2157.67 \\
ModelSmith & 0.305 & 0.997 & 0.785 & 0.795 & 0.589 & 0.643 & 0.673 & 0.711 & 3070.35 \\
Fructose & \textbf{1.000} & \textbf{1.000} & 0.797 & 0.796 & 0.783 & 0.781 & 0.790 & \textbf{0.789} & 951.95 \\
Semantix (Ours) & 0.998 & \textbf{1.000} & 0.804 & \textbf{0.803} & 0.744 & 0.747 & 0.773 & \underline{0.774} & \textbf{730.38} \\
\bottomrule
\end{tabular}
\caption{Comparison of frameworks on the Named Entity Recognition Task with 0 and 2 retries. Metrics include reliability, precision, recall, F1 score, and token usage (reported for 2 retries). The table spans both columns for better readability without overflowing content.}
\label{table:framework_comparison_combined_ner}
\end{table*}

\begin{table*}[!ht]
\scriptsize
\centering
\begin{tabular}{@{}lcc|cc|c@{}}
\toprule
\multirow{2}{*}{Framework} & \multicolumn{2}{c|}{Reliability} & \multicolumn{2}{c|}{Variety} & \multirow{2}{*}{Token Usage} \\
& 0 retries & 2 retries & 0 retries & 2 retries &  \\
\midrule
OpenAI & 0.600 & \textbf{1.000} & \textbf{0.900} & \textbf{0.900} & 224 \\
Instructor & 0.670 & 0.980 & 0.881 & 0.776 & 229.5 \\
LlamaIndex & \textbf{1.000} & 0.960 & 0.030 & 0.021 & \textbf{172} \\
Marvin & \textbf{1.000} & \textbf{1.000} & 0.010 & 0.010 & 616 \\
ModelSmith & 0.790 & \textbf{1.000} & 0.747 & 0.840 & 642.1 \\
Fructose & \textbf{1.000} & \textbf{1.000} & 0.890 & 0.930 & 245.10 \\
Semantix (Ours) & \textbf{1.000} & \textbf{1.000} & 0.820 & 0.820 & \underline{221.48} \\
\bottomrule
\end{tabular}
\caption{Comparison of frameworks on the Synthetic Data Generation task with 0 and 2 retries. Metrics include reliability, variety, and total token usage (reported for 2 retries). The table spans both columns for better readability without overflowing content.}
\label{table:framework_comparison_combined_synthetic_datagen}
\end{table*}

%% file: case_studies.tex
\section{Case Studies}
\subsection{Case Study: CV Analyzer}

In this case study, we explore how Semantix, in collaboration with MTP, addresses the challenge of interpreting Curriculum Vitae (CV) documents to generate robust, structured profiles characterized by nested data types. The objective is to leverage large language models (LLMs) to automate the extraction of relevant information from CVs and facilitate a straightforward job-matching process by comparing the structured profiles with job descriptions.

Semantix's approach abstains from introducing additional abstractions, instead relying on traditional programming constructs. Specifically, Python's built-in data classes are utilized to define cleaner and more maintainable class structures, as shown in the code snippet. The Semantic Type from Semantix is employed to impose meaningful constraints on variables, such as the \texttt{location\_type}, ensuring data consistency and validity.

\begin{minted}[tabsize=2,breaklines, frame=lines,framesep=2mm,, fontsize=\footnotesize]{python}
import semantix as sx

llm = sx.llms.OpenAI()

@dataclass
class Education:
    school_name: str
    degree: str
    field_of_study: str
    start_date: str
    end_date: str
    activities: str
    grade: str
    additional_info: Optional[str] = None

@dataclass
class WorkExperience:
    title: str
    company: str
    location: str
    location_type: sx.Semantic[str, "Onsite | Hybrid | Remote"]
    start_date: str
    end_date: str
    description: str
    additional_info: Optional[str] = None

@dataclass
class Skill:
    name: str
    level: str
    additional_info: Optional[str] = None

@dataclass
class Project:
    name: str
    start_date: str
    end_date: str
    description: str
    url: str
    additional_info: Optional[str] = None

@dataclass
class Certification:
    name: str
    issuing_organization: str
    issue_date: str
    expiration_date: str
    credential_id: str
    additional_info: Optional[str] = None

@dataclass
class Publication:
    title: str
    authors: list[str]
    publication_date: str
    publisher: str
    description: str
    additional_info: Optional[str] = None

@dataclass
class Reference:
    name: str
    position: str
    company: str
    email: str
    phone: str
    additional_info: Optional[str] = None

@dataclass
class Profile:
    name: str
    email: str
    phone: str
    address: str
    summary: str
    education: list[Education]
    experience: list[WorkExperience]
    skills: list[Skill]
    projects: list[Project]
    certifications: list[Certification]
    publications: list[Publication]
    references: list[Reference]
    links: list[str]
    additional_info: list[str] = None

@dataclass
class JobDescription:
    title: str
    company: str
    location: str
    description: str
    requirements: list[str]
    responsibilities: list[str]
    additional_info: Optional[str] = None

@dataclass
class Evaluation:
    summary: str
    education_match: float
    experience_match: float
    skills_match: float
    overall_score: float
\end{minted}

The function enhancement capability of Semantix is harnessed by decorating the functions \texttt{extract\_profile} and \texttt{evaluate\_candidate}. This augmentation allows these functions to utilize a specified LLM to achieve the desired objectives. For the \texttt{evaluate\_candidate} function, the method \texttt{Reason} is incorporated, compelling the LLM to perform a reasoning step before generating structured output, thereby enhancing the quality and reliability of the results.

\begin{minted}[tabsize=2,breaklines, frame=lines,framesep=2mm,fontsize=\footnotesize]{python}
@llm.enhance()
def extract_profile(resume_content: str) -> Profile: ...

@llm.enhance(method="Reason")
def evaluate_candidate(
    profile: Profile, job_description: JobDescription
) -> Evaluation:...
\end{minted}

To evaluate the effectiveness of this approach, various CVs of differing lengths were processed alongside different job descriptions. The system consistently yielded structured outputs, demonstrating its robustness and applicability in real-world scenarios. An example of the generated profile object is presented in Figure~\ref{code:output_cv_analyzer_profile}, which encapsulates detailed candidate information. Similarly, the evaluation object, shown in Figure~\ref{code:output_cv_analyzer_eval}, provides a quantified assessment of the candidate's suitability for a given job, based on factors such as education, experience, and skills match.

\begin{listing*}[!ht]
\centering
\begin{minted}[tabsize=2,breaklines,frame=lines,framesep=2mm,fontsize=\footnotesize]{python}
Profile(name='John Doe', email='john.doe@example.com', phone='123-456-7890', address='123 Main St, Anytown, USA', summary='Aspiring software engineer with a strong background in natural language processing (NLP) and a passion for the impact of AI on human life. Experienced in competitive programming with a proven track record of success in numerous competitions.', education=[Education(school_name='Some High School', degree='High School Diploma', field_of_study='General Studies', start_date='2010-09-01', end_date='2014-06-01', activities='Tennis Team, Scrabble Club', grade='A', additional_info=None), Education(school_name='University of Technology', degree="Bachelor's Degree", field_of_study='Computer Science', start_date='2014-09-01', end_date='2018-06-01', activities='Computer Programming Club, Research in NLP', grade='A', additional_info=None)], experience=[WorkExperience(title='Software Engineer', company='Startup Inc.', location='Remote', location_type='Remote', start_date='2022-01-01', end_date='Present', description='Co-founder and lead developer of a startup focusing on innovative AI applications.', additional_info=None), WorkExperience(title='Software Engineer', company='Amazon', location='Seattle, WA', location_type='Onsite', start_date='2020-01-01', end_date='2021-12-31', description='Worked on various cloud computing projects and contributed to the development of internal tools.', additional_info=None), WorkExperience(title='Intern', company='Google', location='Mountain View, CA', location_type='Onsite', start_date='2018-06-01', end_date='2019-08-31', description='Assisted in NLP research projects and supported software development teams.', additional_info=None)], skills=[Skill(name='Natural Language Processing', level='Advanced', additional_info=None), Skill(name='Python', level='Advanced', additional_info=None), Skill(name='Machine Learning', level='Intermediate', additional_info=None), Skill(name='Competitive Programming', level='Advanced', additional_info=None)], projects=[Project(name='AI Chatbot', start_date='2021-01-01', end_date='2021-06-30', description='Developed a chatbot using NLP techniques to assist users in various tasks.', url='https://github.com/johndoe/ai-chatbot', additional_info=None)], certifications=[Certification(name='Certified Machine Learning Specialist', issuing_organization='AI Institute', issue_date='2020-05-01', expiration_date='2023-05-01', credential_id='ML-123456', additional_info=None)], publications=[Publication(title='The Impact of AI on Human Interaction', authors=['John Doe'], publication_date='2021-09-15', publisher='Tech Journal', description='A research paper discussing the effects of AI technologies on daily human communication and relationships.', additional_info=None)], references=[Reference(name='Jane Smith', position='Manager', company='Startup Inc.', email='jane.smith@example.com', phone='098-765-4321', additional_info=None)], links=['https://linkedin.com/in/johndoe', 'https://github.com/johndoe'], additional_info=['Hobbies include playing tennis, reading sci-fi novels, and participating in hackathons.'])
\end{minted}
\caption{Profile-Object Representing the given Synthetic Curriculum Vitae}
\label{code:output_cv_analyzer_profile}
\end{listing*}

\begin{listing*}[!ht]
\centering
\begin{minted}[tabsize=2,breaklines,frame=lines,framesep=2mm,fontsize=\footnotesize]{python}
Evaluation(summary='John Doe is a strong candidate for the Data Scientist position with relevant education in Computer Science and high proficiency in Python and Machine Learning. However, direct experience in data science roles is somewhat lacking, which affects the experience match.', education_match=100.0, experience_match=70.0, skills_match=85.0, overall_score=85.0)
\end{minted}
\caption{Evaluation-Object Representing the evaluation of candidate profile with respect to the Job Description}
\label{code:output_cv_analyzer_eval}
\end{listing*}

This methodology exemplifies the practical application of LLMs in automating HR processes, reducing manual effort, and improving the accuracy of candidate evaluations. By integrating semantic types and leveraging reasoning capabilities, the system ensures that the extracted data is both meaningful and actionable.

\subsection{Case Study: Food Analyzer}
\label{food-analyzer}

The Food Analyzer case study focuses on utilizing a multimodal LLM to interpret a given food image through Chain-of-Thought reasoning and subsequently provide nutritional information and ingredient lists in a structured format.

The system employs Semantix's LLM capabilities to analyze images, specifically targeting the identification of food items and their nutritional components. By incorporating the Chain-of-Thought methodology, the LLM is guided to think step-by-step, enhancing its ability to reason about the contents of the image and produce more accurate and detailed outputs.

\begin{minted}[tabsize=2,breaklines,frame=lines,framesep=2mm,fontsize=\footnotesize]{python}
from semantix.types import Image

@dataclass
class NutritionInformation:
    calories: int
    protein: int
    carbohydrates: int
    fats: int
    fiber: int
    sodium: int

@dataclass
class FoodAnalysis:
    nutrition_info: NutritionInformation
    ingredients: List[str]
    health_rating: Semantic[str, "How Healthy is the Food"]

@llm.enhance("Analyze the given Food Image", method="CoT")
def analyze(img: Image) ->FoodAnalysis: ...

analysis = analyze(img = Image("examples/ramen.jpg", "high"))
\end{minted}

\begin{figure}[h]
\centering
\includegraphics[width=0.8\linewidth]{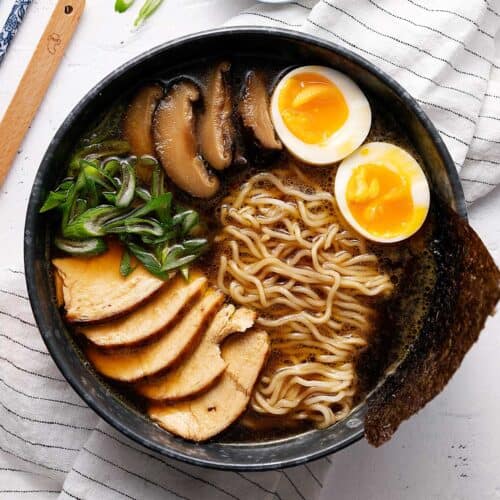} 
\caption{Image of a ramen bowl \href{https://www.elmundoeats.com/quick-30-minute-chicken-ramen/}{(Image Source)}}
\label{img:ramen_bowl}
\end{figure}

An example involves providing the system with an image of a ramen bowl (see Figure~\ref{img:ramen_bowl}). The LLM first identifies the dish and its visible components, such as noodles, broth, and toppings like eggs, vegetables, and meat. It then proceeds to estimate the nutritional content, including calories, protein, carbohydrates, fats, fiber, and sodium levels, based on standard nutritional data for the identified ingredients. The chain-of-thought process is detailed in Output Snippet~\ref{output:food_analyzer_thoughts}.

\begin{listing}
\begin{minted}[tabsize=2,breaklines,frame=lines,framesep=2mm,fontsize=\footnotesize]{python}
FoodAnalysis(nutrition_info = NutritionInformation(calories=600, protein=35, carbohydrates=75, fats=20, fiber=3, sodium=1500), ingredients=['Ramen noodles', 'Sliced chicken', 'Soft-boiled eggs', 'Mushrooms', 'Green onions', 'Seaweed'], health_rating='Moderately Healthy')
\end{minted}
\caption{Structured output of the food analysis.} 
\label{output:food_analyzer_output}
\end{listing}

\begin{listing}
\begin{minted}[tabsize=2,breaklines,frame=lines,framesep=2mm,fontsize=\footnotesize]{markdown}
```chain-of-thoughts
Let's Think Step by Step to achieve the goal.

1. **Identifying the Dish**: The image depicts a bowl of ramen, which typically includes noodles, broth, and various toppings such as eggs, vegetables, and meat.

2. **Ingredients**: The visible ingredients in the dish include ramen noodles, sliced chicken, soft-boiled eggs, mushrooms, green onions, and a seaweed sheet.

3. **Nutritional Analysis**: 
   - **Calories**: A typical bowl of ramen can have around 500-700 calories, depending on the amount of broth, noodles, and toppings used.
   - **Protein**: The chicken and eggs contribute significantly to protein content. Estimated around 30-40 grams.
   - **Carbohydrates**: Primarily from the ramen noodles, estimated around 70-80 grams.
   - **Fats**: Depending on the cooking method and ingredients (like the use of oil), fat content may vary, estimated around 15-25 grams.
   - **Fiber**: Minimal from the vegetables, estimated around 2-5 grams.
   - **Sodium**: Ramen can be high in sodium due to the broth and seasoning, estimated around 1000-2000 mg.

4. **Health Rating**: Ramen can be considered moderately healthy, depending on portion sizes and the balance of ingredients. It provides a good source of protein and carbs but can be high in sodium and fats, so moderation is key.

Now, based on this analysis, we can compile the information into the desired output type.
```
\end{minted}
\caption{Chain of thoughts reasoning process.} 
\label{output:food_analyzer_thoughts}
\end{listing}

The final output is a structured object containing the nutritional information, a list of ingredients, and a health rating, as shown in Output Snippet~\ref{output:food_analyzer_output}. This demonstrates the system's ability to interpret visual data and translate it into meaningful, structured information that can be used for dietary analysis, meal planning, or health assessments.

By integrating multimodal data processing with advanced reasoning techniques, the Food Analyzer showcases the potential of LLMs in fields such as nutrition science, healthcare, and personalized diet recommendations. It highlights how AI can bridge the gap between visual inputs and structured data outputs, providing valuable insights in a user-friendly manner.